\documentclass[runningheads, colorlinks=true, linkcolor=purple, citecolor=purple, urlcolor=purple]{llncs}
\usepackage{graphicx}
\usepackage{multirow}
\usepackage{tikz}
\usetikzlibrary{arrows.meta}
\usepackage{booktabs}
\usepackage{xcolor}
\definecolor{purple}{rgb}{0.5, 0.0, 0.5}
\usepackage{hyperref}
\usepackage[hyphenbreaks]{breakurl}

\usepackage{xspace}

\usepackage[textwidth=0.2\textwidth,textsize=footnotesize
    ,disable
            ]{todonotes}
\usepackage{adjustbox}

\definecolor{PrologPredicate}{RGB}{0,0,200}
\definecolor{PrologVar}      {RGB}{145,032,039}
\definecolor{PrologComment}  {RGB}{169,082,044}
\definecolor{PrologOther}    {rgb}{0.2,0.2,0.2}
\definecolor{PrologString}   {RGB}{070,120,200}
\usepackage{listings} 
\usepackage{wrapfig}
\newcommand{\code}{\lstinline[style=MyInline]}
\lstdefinestyle{MyInline}
{
  basicstyle = \ttfamily\color{PrologOther},
  breaklines = true,
  breakatwhitespace=true,
  upquote = true,
  literate =
  {\ │}{{$\mid$}}1 
  {|}{{$\mid$}}1
  {\\\{}{{\{}}1
  {\\\}}{{\}}}1
  {.=.}{{\#=}}3
  {.<.}{{\#<}}3
  {.>.}{{\#>}}3
  {.=<.}{{\#=<}}4
  {.>=.}{{\#>=}}4
  {<}{{<}}2
  {>}{{>}}2
  {=<}{{=<}}3
  {>=}{{>=}}3
  {\ \\=}{{\,\char"5C=\,}}2
  {\\=}{{\char"5C=}}2
  {?-}{{?-\,}}3
  {:-}{{:-\,}}3
  {\\$}{{\$}}1
}
\lstdefinestyle{Plain}
{
  keywords = {},
  upquote = true,
  basicstyle = \ttfamily\color{PrologPredicate},
  basewidth = 0.48em,
  numbers=none,
  xleftmargin=0cm,
  moredelim = {*[s][\color{black!40!PrologPredicate}]{\#pred}{.}},
  moredelim = {*[s][\color{black!40!PrologPredicate}]{\#show}{.}},
  moredelim = {*[s][\color{black!40!PrologPredicate}]{\#hide}{.}},
  moredelim = {*[s][\color{PrologVar}]{(}{)}},
  moredelim = {*[s][\color{PrologString}]{'}{'}},
 commentstyle = \mdseries\color{PrologComment},
  morecomment=[l]\%,
   literate     =
  {|}{{$\mid$}}1
  {\\$}{{\$}}1
  {\ │}{{$\mid$}}1,
}
\lstdefinestyle{MySCASP}
{
  keywords = {},
  upquote = true,
  basicstyle = \ttfamily\color{PrologPredicate},
  basewidth = 0.48em,
  moredelim = {**[is][\color{PrologComment}]{`}{`}},
  moredelim = {*[s][\color{black!40!PrologPredicate}]{\#pred}{.}},
  moredelim = {*[s][\color{black!40!PrologPredicate}]{\#show}{.}},
  moredelim = {*[s][\color{black!40!PrologPredicate}]{\#hide}{.}},
  moredelim = {*[s][\color{PrologVar}]{(}{)}},
  moredelim = {*[s][\color{PrologString}]{'}{'}},
  moredelim = {*[s][\color{PrologOther}]{:-}{.}},
  moredelim = {*[s][\color{red}]{/*}{*/}},
  commentstyle = \mdseries\color{PrologComment},
  morecomment=[l]\%,
  morecomment=[s]{/*}{*/},
  literate     =
  {|}{{$\mid$}}1
  {\ │}{{$\mid$}}1
  {[}{{\color{PrologOther}\small[}}1
  {]}{{\color{PrologOther}\small]}}1
  {\\$}{{\$}}1
  {&(}{{\color{PrologOther}(}}1
  {&)}{{\color{PrologOther})}}1
  {&.}{{.}}0
  {\\=}{{\char"5C=}}2
  {\\$}{{\$}}1,
}
\newcommand{\myurl}[1]{\href{http://platon.etsii.urjc.es/~jarias/papers/dcc-padl21/#1}{\nolinkurl{#1}}}

\begin{document}
\title{Towards Dynamic Consistency Checking in 
    \mbox{Goal-directed Predicate Answer Set Programming}
}
\titlerunning{DCC in s(CASP)}
%
\author{%
  Joaqu\'{\i}n Arias\inst{1}\orcidID{0000-0003-4148-311X} \and
  Manuel Carro\inst{2,3}\orcidID{0000-0001-5199-3135} \and
  Gopal Gupta\inst{4}\orcidID{0000-0001-9727-0362}
}
\authorrunning{J. Arias et al.}
%

\institute{%
  CETINIA, Universidad Rey Juan Carlos, Madrid, Spain \and%
  Universidad Polit\'ecnica de Madrid, Spain \and
  IMDEA Software Institute, Madrid, Spain \and
  University of Texas at Dallas, Richardson, USA\\
  \email{joaquin.arias@urjc.es},
  \email{manuel.carro@\{upm.es,imdea.org\}},
  \email{gupta@utdallas.edu}%
}
\maketitle              
\begin{abstract}

  \hyphenpenalty 5000
  
  Goal-directed evaluation of Answer Set Programs is gaining traction
  thanks to its amenability to create AI systems that can, due to the
  evaluation mechanism used, generate explanations and justifications.
  s(CASP) is one of these systems and has been already used to write
  reasoning systems in several fields.  It provides enhanced
  expressiveness w.r.t.\ other ASP systems due to its ability to use
  constraints, data structures, and unbound variables natively.
  However, the performance of existing s(CASP) implementations is not
  on par with other ASP systems: model consistency is checked once
  models have been generated, in keeping with the
  generate-and-test paradigm.
  In this work, we present a variation of the top-down evaluation
  strategy, termed \emph{Dynamic Consistency Checking}, which
  interleaves model generation and consistency checking. This makes it
  possible to determine when a literal is not compatible with the
  \emph{denials} associated to the global constraints in the program,
  prune the current execution branch, and choose a different
  alternative.  This strategy is specially (but not exclusively)
  relevant in problems with a high combinatorial component.  We have
  experimentally observed speedups of up to 90$\times$ w.r.t.\ the
  standard versions of s(CASP).
  
\end{abstract}

\section{Introduction}

s(CASP)~\cite{scasp-iclp2018} is a novel non-monotonic reasoner that
evaluates Constraint Answer Set Programs without a grounding phase,
either before or during execution.
s(CASP) supports predicates and thus retains logical
variables (and constraints) both during the execution and in the
answer sets.
The operational semantics of s(CASP) relies on backward chaining,
which is intuitive to follow and lends itself to generating
explanations that can be translated into natural
language~\cite{arias-TC-explain-scasp}. The execution of an s(CASP)
program returns partial stable models: the subsets
of the stable models~\cite{gelfond88:stable_models} which include
only the (negated) literals necessary to support the initial query. To
the best of our knowledge, s(CASP) is the only system that exhibits
the property of relevance~\cite{pereira89:relevant-counterfactuals}.
s(CASP) has been already applied in relevant fields related
to the representation of commonsense reasoning:

\begin{itemize}
\item An automated reasoner that uses Event Calculus
  (EC)~\cite{arias19:event-calculus-asp-lopstr19}
  (\url{http://bit.ly/EventCalculus}).  s(CASP) can make deductive
  reasoning tasks in domains featuring constraints involving dense
  time and continuous properties. It is being used to model real-world
  avionics systems, to verify (timed) properties, and to identify gaps
  with respect to system requirements~\cite{hall2021knowledge}.
    
\item The s(CASP) justification framework has been used to bring
  Explainable Artificial Intelligence (XAI) principles to rule-based
  systems capturing expert
  knowledge~\cite{arias-TC-explain-scasp,chen2016physician}.

\item It is at the core of two natural language understanding
  systems~\cite{basu2021knowledge}: SQuARE, a Semantic-based Question
  Answering and Reasoning Engine, and StaCACK, Stateful Conversational
  Agent using Commonsense Knowledge.  They use the s(CASP) engine to
  ``truly understand'' and perform reasoning while generating 
  natural language explanations for their responses. Building on these
  systems, s(CASP) was used to develop one of the nine systems
  selected to participate in the Amazon Alexa Socialbot Grand
  Challenge 4
  \cite{alexa},\footnote{\url{https://cs.utdallas.edu/computer-scientists-enhance-alexas-small-talk-skills/}}
  and is being used to develop a conversational AI chatbot.
  
\item It has been used in ILP systems that generate ASP
  programs~\cite{DBLP:conf/aaai/ShakerinG19-short} and concurrent
  imperative programs from behavioral, observable
  specifications~\cite{DBLP:conf/lopstr/VaranasiSMG19}.

\item A legal expert system~\cite{DBLP:conf/icail/Morris21}, developed
  at the SMU Centre for Computational Law at Singapore, coded rule 34
  of Singapore's Legal
  Profession.\footnote{\url{https://github.com/smucclaw/r34_sCASP}}
  Its front-end is a web interface that collects user information,
  runs queries on s(CASP), and displays the results with explanations
  in natural language.
  
\item s(LAW), an administrative and judicial discretion
  reasoner~\cite{ariaslaw}, which allows modeling legal rules involving
  ambiguity and infers conclusions, providing (natural language)
  justifications based on them.
  
\end{itemize}

However, in the standard implementation of s(CASP),
the global constraints in a program are checked
when a tentative but complete model is computed. This
strategy takes a large toll on the performance of programs that
generate many tentative models and use global constraints to discard
those that do not satisfy the specifications of the problem.

In this work, we propose a technique termed \emph{Dynamic Consistency
  Checking} (DCC) that anticipates the evaluation of global
constraints to discard inconsistent models as early as possible.
Before adding a literal to the tentative model, DCC checks if any
global constraint is violated. If so, this literal is discarded and
the evaluation backtracks to look for other alternatives.  We show,
through several examples, that using this preliminary implementation,
s(CASP) with DCC is up to 90$\times$ faster.
Section~\ref{sec:syntax-scasp} contains an overview of the syntax,
operational semantics, and implementation of s(CASP).
Section~\ref{sec:dcc} explains the motivation behind DCC with
examples, and describes its design and  implementation
Section~\ref{sec:evaluation} presents the evaluating results of
several benchmarks using s(CASP) with DCC 
enabled or not. Finally, in Section~\ref{sec:conclusions} we draw
conclusions and propose future work.
All the program and files used or mentioned in this paper are
available at
\url{http://platon.etsii.urjc.es/~jarias/papers/dcc-padl21/}.

\section{Background: s(CASP)}
\label{sec:syntax-scasp}

An s(CASP) program is a set of clauses of the following form:
\smallskip

{\centering \code{a:-c$_a$, b$_1$, $\dots$, b$_m$, not b$_{m+1}$, $\dots$, not b$_n$.} \par}

\smallskip

\noindent where \code{a} and \code{b$_1$, $\dots$, b$_n$} are atoms.
An atom is either a propositional variable or the expression
\code{p(t$_1$, $\ldots$, t$_n$)} if \code{p} is an $n$-ary predicate
symbol and \code{t$_1$, $\ldots$, t$_n$} are terms.
A term is either a variable \code{x$_i$} or a function symbol \code{f}
of arity \code{n}, denoted as \code{f/n}, applied to $n$ terms, e.g.,
\code{f(t$_1$, t$_2$, $\ldots$, t$_n$)}, where each \code{t$_i$} is in
turn a term.  A function symbol of arity 0 is called a constant.
Therefore, s(CASP) accepts terms with the same conventions as Prolog:
\code{f(a, b)} is a term, and so are \code{f(g(X),Y)} and
\code{[f(a)|Rest]} (to denote a list with head \code{f(a)} and tail
\code{Rest}).  Program variables are usually written starting with an
uppercase letter,\footnote{There are additional syntactical
  conventions to distinguish variables and non-variables that are of no
  interest here.} while function and predicate symbols start with a
lowercase letter.  Numerical constants are written solely with digits.

The term \code{c$_a$} is a simple constraint or a conjunction of
constraints: an expression establishing relations among variables in
some constraint system~\cite{intro_constraints_stuckey}.  Similar to
CLP, s(CASP) is parametrized w.r.t.\ the constraint system, from which
it inherits its semantics.  Since the execution of an s(CASP) program
needs negating constraints, soundness requires that this can be done
in the chosen constraint system by means of a finite disjunction of
basic constraints~\cite{dover2000:constructive-negation,Stuckey91}.

At least one of \code{a}, \code{b$_i$}, or \code{not b$_i$} must be
present. When the head \texttt{a} is not present, it is supposed to be
substituted by the head \emph{false}.  Headless rules
have then the form \smallskip

{\centering
\code{:-c$_a$, b$_1$, $\dots$, b$_m$, not b$_{m+1}$, $\dots$, not b$_n$.}   
\par}

\medskip

\noindent
and their interpretation is that the conjunction of the constraints
and goals has to be false, so at least one constraint or goal has to
be false.
For example, the rule \code{:-p, q}, expresses that the conjunction of
atoms %
\code{p $\land$ q} cannot be true: either \code{p}, \code{q}, or both,
have to be false in any stable model.
ASP literature often uses the term \emph{constraint} to denote these
constructions.
To avoid the ambiguity that may arise from using the same name for
constraints appearing among (free) variables during program execution
and in the final models and for rules without heads, we will refer to
headless rules as \emph{denials}~\cite{mellarkod-long}.

The execution of an s(CASP) program starts with a \emph{query} of the
form %
\smallskip

{\centering \code|?- c$_a$, b$_1$, $\dots$, b$_m$, not b$_{m+1}$, $\dots$, not b$_n$.|  \par}

\medskip

The s(CASP) answers to a query are \emph{partial} stable models where
each one is a subset of a stable model that satisfies the constraints,
makes non-negated atoms true, makes the negated atoms non-provable,
and, in addition, includes only atoms that are relevant to support the
query. Additionally, for each partial stable model s(CASP) returns on
backtracking \emph{partial answer sets} with the justification tree
and the bindings for the free variables of the query that correspond
to the most general unifier (\emph{mgu}) of a successful top-down
derivation.

\subsection{Execution Procedure of s(CASP)}
\label{sec:overview}

Let us present an abridged description of the top-down strategy of
s(CASP):

\begin{enumerate}

\item\label{item:consneg} Rules expressing the constructive negation
  of the predicates in the original ASP program are synthesized.  We
  call this the \emph{dual program}.  Its mission is to provide a
  means to constructively determine the conditions and constraints
  under which calls to non-propositional predicates featuring
  variables would have failed: if we want to know when a rule such as
  \code{p(X,Y):-q(X), not r(Y)} succeeds, the dual program computes
  the constraints on \code{Y} under which the call \code{r(Y)} would
  fail.  This is an extension of the usual ASP semantics that is
  compatible with the case of programs that can be finitely
  grounded.\footnote{Note that, in the presence of function symbols
    and constraints on dense domains, this is in general not the case
    for s(CASP) programs.}
  A description of the construction of the dual program can be found
  in~\cite{arias21:event-calculus-asp-arxiv,scasp-iclp2018,marple2017computing}.

\item\label{item:negloops} The original program is checked for loops
  of the form \code{r:-q, not r.} and introduces additional denials to
  ensure that the models satisfy $\lnot q \lor r$, even if the atoms
  \code{r} or \code{q} are not needed to solve the query.
  This is done by building a dependency graph of the program and
  detecting the paths where this may happen, including call paths
  across calls.  Therefore, for the program:

  \begin{multicols}{3}
\begin{lstlisting}[style=MySCASP]
p :- not q.
q :- not p.
r :- not r.
\end{lstlisting}
  \end{multicols}
  
  \noindent
  s(CASP) will determine that there are no stable models, regardless
  of the initial query.
  For the propositional case, such an analysis can be precise.  For
  the non-propositional case, an over-approximation is calculated.  In
  both cases, denials that are not used during program evaluation can
  be generated.  These may impose a penalty in execution time, but are
  safe.
  
\item\label{item:nmr-intro} The denials generated in point~\ref{item:negloops}, together
  with any denials present in the original program, are collected in
  predicates synthesized by the compiler that are evaluated by adding
  an auxiliary goal, \code{nmr_check/0}, at the end of the top-level
  query.

\item\label{item:interp} The union of the original program, the dual
  program, and the denials is handled by a top-down %
  algorithm that implements the stable model semantics.
\end{enumerate}

Point number~\ref{item:interp} is specially relevant.  The dual
program (point~\ref{item:consneg}) is synthesized by means of program
transformations drawing from classical logic.  However, its intended
meaning differs from that of first-order logic.  That is so because it
is to be executed by a metainterpreter that does not implement the
inference mechanisms of first-order logic, as it is designed to ensure
that the semantics of answer set programs is respected (see
Section~\ref{sec:scasp-algorithm}).  In particular, it treats
specifically cyclic dependencies involving negation.

\subsection{Unsafe Variables and Uninterpreted Function Symbols}
\label{sec:unsafe}

The following code, from~\cite[Pag.~9]{arias21:event-calculus-asp-arxiv}
has variables that would be termed as \emph{unsafe} in regular ASP
systems: variables that appear in negated atoms in the body of a
clause, but that do not appear in any positive literal in the same
body.

    \begin{multicols}{2}
\begin{lstlisting}[style=MySCASP]
p(X):- q(X, Z), not r(X).  
p(Z):- not q(X, Z), r(X).
q(X, a):- X #> 5.  
r(X):- X #< 1.
\end{lstlisting}
    \end{multicols}

Since s(CASP) synthesizes explicit constructive goals for these
negated goals, the aforementioned code can be run as-is in s(CASP).
The query %
\code{?-p(A).} generates three different answer sets, one for each
binding:

\begin{lstlisting}[style=MySCASP, numbers=none, xleftmargin=0em]
{ p(A| {A #> 5}), q(A| {A #> 5}, a), not r(A| {A #> 5}) }
  A #> 5
{ p(A| {A \= a}), not q(B| {B #< 1}, A| {A \= a}), r(B| {B #< 1}) } 
  A \= a
{ p(a), not q(B| {B #< 1}, a), r(B| {B #< 1}) } 
  A = a
\end{lstlisting}

\noindent
where the notation \code'V|{C}' for a variable \code{V} is intended
to mean that \code{V} is subject to the constraints in \code|{C}|.
The constraints \code{A = 5}, \code{A $\not=$ a}, and \code{A = a}
correspond to the bindings of variable \code{A} that make the atom
from the query \code{?-p(A)}  belong to the stable model.

Another very relevant point where s(CASP) differs from ASP is in the
possibility of using arbitrary uninterpreted function symbols to
build, for example, data structures.  While in mainstream ASP
implementations these could give rise to an infinite grounded program
(i.e., if the program does not have the \emph{bound-term-depth} property),
the s(CASP) execution model can deal with them similarly to Prolog, with
the added power of the use of constructive negation in the execution
and in the returned models.

\begin{example}
  \label{ex:using-lists}
  The predicate \code{member/2} below,
  from~\cite[Pag.~11]{arias21:event-calculus-asp-arxiv}, models the
  membership to a list as it is usual in (classical) logic
  programming.  The query is intended to derive the conditions for one
  argument not to belong to a given list.

  \begin{multicols}{2}
\begin{lstlisting}[style=MySCASP, basewidth=.45em]
member(X, [X|Xs]).  
member(X, [_|Xs]):- member(X, Xs).
list([1,2,3,4,5]).
?- list(A), not member(B, A).
\end{lstlisting}
  \end{multicols}
  
\noindent
This program and query return in s(CASP) the following model and
bindings:

\begin{lstlisting}[style=Plain, numbers=none, xleftmargin=0em]
{ list([1,2,3,4,5]),
   not member(B| {B \= 1,B \= 2,B \= 3,B \= 4,B \= 5}, [1,2,3,4,5]), 
   not member(B| {B \= 1,B \= 2,B \= 3,B \= 4,B \= 5}, [2,3,4,5]),
   not member(B| {B \= 1,B \= 2,B \= 3,B \= 4,B \= 5}, [3,4,5]),
   not member(B| {B \= 1,B \= 2,B \= 3,B \= 4,B \= 5}, [4,5]), 
   not member(B| {B \= 1,B \= 2,B \= 3,B \= 4,B \= 5}, [5]),
   not member(B| {B \= 1,B \= 2,B \= 3,B \= 4,B \= 5}, []) }
   A = [1,2,3,4,5], B \= 1, B \= 2, B \= 3, B \= 4, B \= 5
\end{lstlisting}
\end{example}

\noindent
I.e., for variable \code{B} not to be a member of the list
\code{[1,2,3,4,5]} it has to be different from each of its elements.

\subsection{s(CASP) as a Conservative Extension of ASP}
\label{sec:semantics}

The behavior of s(CASP) and ASP is the same for propositional
programs.
They differ in programs with unsafe variables (not legal in mainstream
ASP systems), programs that could create unbound data structures, or
whose variable ranges are defined in infinite domains (either unbound
or bound but dense).  Such programs are outside the standard domain of
ASP systems as they cannot be finitely grounded, For them, s(CASP)
extends ASP consistently.

In addition, the domain of the variables is implicitly expanded to
include a domain which can be potentially infinite.  Let us use the
following example,
from~\cite[Pag.~12]{arias21:event-calculus-asp-arxiv}, where we are
interested in knowing whether \code{p(X)} (for some \code{X}) is or
not part of a stable model:
 
\begin{multicols}{2}
\begin{lstlisting}[style=MySCASP]
d(1).
p(X) :- not d(X).
\end{lstlisting}
\end{multicols}

The only constant in the program is \code{1}, which is the only
possible domain for \code{X} in the second clause.  That clause is not
legal for ASP, as \code{X} is an unsafe variable (Section~\ref{sec:unsafe}).
Adding a domain
predicate call for it (i.e., adding \code{d(X)} to the body of the
second clause), makes its model be \code|{d(1)}| (\code{not p(1)}
is implicit).

That second clause is however legal in s(CASP).  Making the query
\code{?-p(X)} returns the \emph{partial} model %
\code'{p(X| {X \= 1}), not d(X| {X \= 1})}' stating that \code{p(X)} and
\code{not d(X)} are true when %
\code{X \= 1}, which is consistent with, but more general than, the
model given by ASP.  As the model is partial, only the atoms (perhaps negated)
involved in the proof for \code{?-p(X)} appear in that model.

\subsection{The s(CASP) Interpreter}
\label{sec:scasp-algorithm}

\begin{figure}[tb]
\begin{multicols}{2}
\begin{lstlisting}[style=MySCASP, basewidth=.5em, escapechar=@]
scasp(Query) :-
   solve(Query,[],Mid),      @\label{scasp:q}@
   solve_goal(nmr_check,Mid,Model),    @\label{scasp:nmr}@
   print_just_model(Model).       @\label{scasp:just}@

solve([],In,['\$success'|In]).
solve([Goal|Gs],In,Out) :-
   solve_goal(Goal,In,Mid),
   solve(Gs,Mid,Out).

solve_goal(Goal,In,Out) :-
   user_defined(Goal), !,     @\label{scasp:user}@
   check_loops(Goal,In,Out).
solve_goal(Goal,In,Out) :-
   Goal=forall(Var,G), !,     @\label{scasp:forall}@
   c_forall(V,G,In,Out).      @\label{scasp:c_f}@
solve_goal(Goal,In,Out) :-
   call(Goal),                @\label{scasp:call}@
   Out=['\$success',Goal|In].

check_loops(Goal,In Out) :-
   loop_type(Goal,In,Loop),
   s_loop(Loop,Goal,In,Out).

s_loop(odd,_,_,_) :- fail.       @\label{scasp:odd}@
s_loop(even,G,In,[chs(G)|In]).   @\label{scasp:even}@
s_loop(no_loop,G,In,Out) :-      @\label{scasp:noloop}@
   pr_rule(G, Body),
   solve(Body,[G|In],Out).
s_loop(proved,G,In,[proved(G)|In]).
s_loop(positive,_,_,_) :- fail.
\end{lstlisting}
  \end{multicols}
  \caption{Outline of the s(CASP) interpreter's code implemented in
    Ciao Prolog.}
\label{fig:interpreter}
\end{figure}

Queries to the original program extended with the dual rules are
evaluated by a runtime environment.  This is currently a
metainterpreter (see Fig.~\ref{fig:interpreter}) in Prolog that
executes an algorithm~\cite{marple2012:goal-directed-asp} that has
similarities with SLD resolution. But it takes into account specific
characteristics of ASP and the dual programs, such as the denials, the
different kinds of loops, and the introduction of universal
quantifiers in the body of the clauses:

\begin{enumerate}

\item The query \code{Query} is evaluated invoking \code{solve/3} in
  line~\ref{scasp:q} starting with an empty model represented as  the empty
  list \code{[]}.
\item \label{item:denial} After the query evaluation, and to ensure
  that the returned model,
  \code{Mid}, is consistent with the denials, \code{nmr_check}
  (item~\ref{scasp:nmr} in page~\pageref{scasp:nmr}) is evaluated in
  line~\ref{scasp:nmr}.
  \item In line~\ref{scasp:just}, the models that are consistent (and
    their justifications), \code{Model}, are output by
    \code{print_just_model/1}.

  \item The predicate \code{solve/3} receives a list with the literals
    in the query (or in the body of some rule) and evaluates them, one
    by one, invoking \code{solve_goal/3}.
  
\item When the literal is a user defined predicate
  (line~\ref{scasp:user}), the interpreter checks if there is a loop
  invoking \code{check_loops/3}.  Three main cases are distinguished
  by \code{type_loop/3}:

  \begin{description}
  \item [Odd loop] When a call eventually invokes itself and there is
    an odd number of intervening negations (as in, e.g., %
    \quad \code{p:- q. $~$ q:- not r. $~$ r:- p.}), the evaluation
    fails in line~\ref{scasp:odd}, to avoid contradictions of the form
    $p \land \lnot p$, and backtracking takes place.
  \item [Even loop] When there is an even number of intervening
    negations (as in %
    \code{p:- not q. $~$ q:- r. $~$ r:- not p.}),  the metainterpreter
    succeeds in line~\ref{scasp:even} to generate several stable
    models, such as %
    \code|{p, not q, not r}| and \code|{q, r, not p}|.
  \item [No loop] If no loops are detected, in line~\ref{scasp:noloop}
    the interpreter invokes \code{pr_rule/2} to retrieve the rule that
    unifies with the goal \code{G} and continues the evaluation by
    invoking \code{solve/3} with the literals of the rule.
  \end{description}

\item The construction \code{forall(Var, G)} in
  line~\ref{scasp:forall} is the dual of the existential
  quantifications in the body of the clauses. To evaluate them the
  runtime environment invokes the predicate \code{c_forall/4} in
  line~\ref{scasp:c_f}, which determines if \code{G} holds for all the
  values of \code{Var} --- see~\cite{scasp-iclp2018} for
  implementation details.
  
\item Finally, operations involving constraints and/or builtins are
  natively handled by invoking \code{call/1} in
  line~\ref{scasp:call}.

\end{enumerate}

\section{Dynamic Consistency Checking in s(CASP)}
\label{sec:dcc}

The Dynamic Consistency Checking proposal of~\cite{marple14:dynamic}
is designed for propositional programs, while our proposal can also
take care of predicate ASP programs.  It is based on anticipating the
evaluation of denials to fail as early as possible

\subsection{Motivation}

As we mentioned before, a denial such as \code{:-p, q}, expresses that
the conjunction of atoms %
\code{p $\land$ q} cannot be true: either \code{p}, \code{q}, or both,
have to be false in any stable model.
In predicate ASP the atoms have variables and a denial such as
\code{:-p(X), q(X,Y)} means that:
$$false \ \ \leftarrow \ \exists x,y\ (\  p(x) \land q(x,y)\ ) $$

\noindent
i.e., \code{p(X)} and \code{q(X,Y)} can not be simultaneously true for
any possible values of \code{X} and \code{Y} in any stable
model.  To ensure that the tentative partial model is
consistent with this denial, the compiler generates a rule of the
form
$$\forall x,y \ ( chk_i \ \ \leftrightarrow \ \lnot (\  p(x) \land q(x,y)\ )\ ) $$

\noindent
and to ensure that each sub-check ($chk_i$) is satisfied, they are
included in the rule
$nmr\_check \leftarrow chk_1 \land \dots \land chk_k \land \ldots$,
which is transparently called after the program query by the s(CASP)
interpreter (see Fig.~\ref{fig:interpreter}, line~\ref{scasp:nmr}).

However, this generate-and-test strategy has a high impact on the
performance of programs that create many tentative models and use
denials to discard those that do not satisfy the constraints of the
problem.

\begin{figure}[tb]
\begin{lstlisting}[style=MySCASP, escapechar=@]
reachable(V) :- chosen(a, V).                @\label{ham:r1}@
reachable(V) :- chosen(U, V), reachable(U).  @\label{ham:r2}@
chosen(U, V) :- edge(U, V), not other(U, V).     % Choose or not an  @\label{ham:chosen}@
other(U, V) :- edge(U, V), not chosen(U, V).     % edge of the graph.

:- vertex(U), not reachable(U).      % Every vertex must be reachable. @\label{ham:d1}@
:- chosen(U, W), U \= V, chosen(V, W).     % Do not choose edges to or  @\label{ham:d2}@
:- chosen(W, U), U \= V, chosen(W, V).     % from the same vertex.     @\label{ham:d3}@
#show chosen/2.

?- reachable(a).                      % Is there a path from a to a?   @\label{ham:q}@
\end{lstlisting}
    \caption{Code of the Hamilonian problem à la ASP, available at
      \myurl{hamiltonian.pl}.}
    \label{fig:hamiltonian}
  \end{figure}

  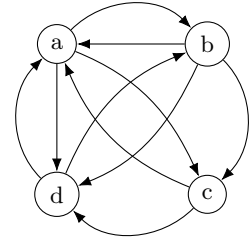
\begin{figure}[tb]
  \begin{minipage}{.65\linewidth}
    \begin{multicols}{3}
\begin{lstlisting}[style=MySCASP]
% Graph
vertex(a).
vertex(b).
vertex(c).
vertex(d).
edge(b, a).
edge(b, d).
edge(a, c).
edge(a, b).
edge(b, c).
edge(c, d).
edge(d, a).
edge(c, a).
edge(a, d).
edge(d, b).
\end{lstlisting}
    \end{multicols}
  \caption{Graph with 4 nodes available at \myurl{graph_4.pl}.}
  \label{fig:graph}
  \end{minipage}
  \begin{minipage}{.04\linewidth}
    ~
  \end{minipage}
  \begin{minipage}{.29\linewidth}
      \begin{tikzpicture}[node distance=2cm, transform shape]
        \node (a) [circle, draw=black] {a};
        \node (b) [circle, draw=black, right of=a] {b};
        \node (d) [circle, draw=black, below of=a] {d};
        \node (c) [circle, draw=black, right of=d] {c};

        \draw [draw, -{Latex[length=1.75mm]}] (a) to[out=45, in=135] (b);
        \draw [draw, -{Latex[length=1.75mm]}] (b) to[out=-45, in=45] (c);
        \draw [draw, -{Latex[length=1.75mm]}] (c) to[out=225, in=-45] (d);
        \draw [draw, -{Latex[length=1.75mm]}] (d) to[out=135, in=225] (a);

        \draw [draw, -{Latex[length=1.75mm]}] (a) to[out=-22, in=115] (c);
        \draw [draw, -{Latex[length=1.75mm]}] (d) to[out=67, in=202] (b);
        \draw [draw, -{Latex[length=1.75mm]}] (b) to[out=180, in=0] (a);

        \draw [draw, -{Latex[length=1.75mm]}] (a) to[out=-90, in=90] (d);
        \draw [draw, -{Latex[length=1.75mm]}] (b) to[out=-115, in=22] (d);
        \draw [draw, -{Latex[length=1.75mm]}] (c) to[out=158, in=-67] (a);

      \end{tikzpicture}    
  \end{minipage}
\end{figure}

\begin{example}[Hamiltonian path problem]\label{exa:hamiltonian}

  Consider the Hamiltonian path problem, in which for a given graph we
  search for a cyclic path that visits each node of the graph only
  once.
  The standard ASP code for this problem, available at
  \myurl{hamiltonian.pl}, is in Fig.~\ref{fig:hamiltonian}.
  The conditions of the problem are captured (i) in line~\ref{ham:d1}
  to discard tentative paths that do not visit all the nodes, and (ii)
  in lines~\ref{ham:d2}-\ref{ham:d3} to discard paths that have edges
  violating the properties of the Hamiltonian path.
  For the query in line~\ref{ham:q}, using the graph in
  Fig.~\ref{fig:graph} there are three stable models, one for each
  Hamiltonian cycle:

\begin{lstlisting}[style=Plain]
{ chosen(a,c),  chosen(c,d),  chosen(d,b),  chosen(b,a),$\dots$ }
{ chosen(a,b),  chosen(b,c),  chosen(c,d),  chosen(d,a),$\dots$ }
{ chosen(a,d),  chosen(d,b),  chosen(b,c),  chosen(c,a),$\dots$ }
\end{lstlisting}
  
As mentioned before, the standard s(CASP) execution follows a
generate-and-test scheme, choosing a cycle that reaches node \code{a}
from node \code{a} and then discards any cycle in which:

  \begin{itemize}
  \item Not all vertices are reached (line~\ref{ham:d1}), e.g.,
    \code{\{chosen(a,b),chosen(b,a)\}}.
  \item Two chosen edges reach / leave the same vertex
    (line~\ref{ham:d2}), e.g., %
    \code{\{chosen(a,b), chosen(d,b), chosen(b,a)\}} or
    \code{\{chosen(a,b), chosen(b,d), chosen(b,a)\}}.
  \end{itemize}

  As a consequence, if the evaluation chooses an edge that breaks any
  of these conditions, trying combinations with the rest of the
  edges would be misused effort.
\end{example}

\subsection{Outline of the DCC Approach}

The main idea behind our proposal is to anticipate the evaluation of
the denials to fail and backtrack as early as possible.  When an atom
involved in a denial is a candidate to be added to a tentative model,
the denial is checked to ensure that it is not already violated.  By
contrast, the classical implementation of s(CASP) checked the
tentative model as a whole once it had been completely built.  The
latter is akin to the generate-and-test approach to combinatorial
problems, while the former tries to cut the search early.

In the most general case, DCC can take the form of 
a constraint-and-test instance.  While detecting some denial
violations can be performed by just checking the current state of the
(partial) candidate model, better, more powerful algorithms can impose
additional constraints on the rest of the atoms belonging to the
partial model and also on the parts of the denials that remain to be
checked.

These constraints can propagate additional conditions through the
candidate model to ensure that it is consistent with the denials.
These conditions will also remain active for the rest of the
construction of the model, so that they can be carried forward,
further reducing the search space.  Note that since s(CASP) includes
constraints \emph{à la} CLP, the effect is very similar to the
constraint propagation mechanisms that take place in constraint
satisfaction systems, therefore making a full s(CASP) + DCC system an
instance of a constraint-and-generate evaluation engine.

The current implementation, which we describe in this paper, is a
\emph{proof of concept} that only checks grounds literals
and does not anticipate the consistency check of constrained literals.
As we will see later, this does not have a negative impact on the
soundness of the system.

\subsection{Implementation of DCC in s(CASP)}

The implementation of s(CASP) + DCC is available as part of the source
code of s(CASP) at
\url{https://gitlab.software.imdea.org/ciao-lang/scasp}.

\subsubsection{Compilation of the Denials}
\label{sec:compilation-denials}

As we mentioned before, the denials are compiled in such a way that
the interpreter checks consistency by proving that forall possible
values, the negation of the denial is satisfied.
For example, for the denial \code{:-p(X), q(X,Y)}, the compiler
generates the rule:

\begin{lstlisting}[style=MySCASP]
chk1 :- forall(X, forall(Y, not chk_body(X,Y))).
not chk_body(X,Y) :- not p(X).
not chk_body(X,Y) :- p(X), not q(X,Y).
\end{lstlisting}

The last clause includes a call to \code{p(X)} to avoid duplicated
solutions provided by the two clauses.
If the interpreter is able to prove that for all the possible values
of \code{X} and \code{Y} the tentative partial model is consistent
with the predicate \code{not chk_body(X,Y)}, then it means that the
tentative partial model is a stable partial model.

The approach followed by the DCC proposal is to detect when a rule
like the above determines that a model candidate is inconsistent.
If that is the case, s(CASP) fails and provokes backtracking to
explore the generation of a different model.  In the example above, if
a model being generated is consistent with \code{p(X) $\land$ q(X,Y)},
then it should be discarded.

Since it is only necessary to check for violation of denials when
adding a goal involved in one of the denials,
the compiler creates a series of rules that state what has to be
checked for in case a goal involved in the denial is generated.  For
the case above, if \code{p(X)} is added, then \code{q(X,Y)} has to be
checked to ensure it does not hold, and the other way around.  This is
represented as:

\begin{lstlisting}[style=MySCASP]
dcc(p(X), [q(X,Y)]).
dcc(q(X,Y), [p(X)]).
\end{lstlisting}
which can be understood as ``if \code{p(X)} is present, check that
\code{q(X,Y)} is not present''.

In general, given a denial of the form %
\code{:-c$_a$, b$_1$, $\dots$, b$_n$} %
for each (negated) literal \code{b$_k$}, of a user defined predicate,
the compiler generates a DCC rule %
\code{dcc(b$_k$, [c$_a$, $\dots$, b$_{k-1}$, b$_{k+1}$, $\dots$])} for
each $k, 1 \leq k \leq n$.

\begin{figure}[tb]
\begin{multicols}{2}
\begin{lstlisting}[style=MySCASP, basewidth=.5em, escapechar=@]
% The only clauses that changes
s_loop(even,G,In,[chs(G)|In]) :-
   eval_dcc(G,In).   % New call
s_loop(no_loop,G,In,Out) :-
   pr_rule(G, Body),
   solve(Body,[G|In],Out),
   eval_dcc(G,In).   % New call

eval_dcc(G,In):-
   \+ &( ground(G),                   @\label{dcc:ground}@
        pr_dcc_rule(G,F_Atoms),      @\label{dcc:dcc}@
        holds_dcc(F_Atoms,In) &).    @\label{dcc:holds}@
holds_dcc([],_).
holds_dcc([F_A|F_As],In) :-
   holds_dcc_(F_A,In),
   holds_dcc(F_As,In).


holds_dcc_(F_A,In) :-
   user_defined(F_A), !,
   member(F_A,In).          @\label{dcc:member}@
holds_dcc_(F_A,In) :-
   call(F_A).               @\label{dcc:call}@

\end{lstlisting}
  \end{multicols}
  \caption{Outline of the changes to the s(CASP) interpreter extended with DCC.}
  \label{fig:mod}
\end{figure}

\subsubsection{Extending the s(CASP) interpreter with DCC}

Fig.~\ref{fig:mod} shows the relevant fragment of the s(CASP)
interpreter extended with Dynamic Consistency Checking.  The basic
intuition is that as soon as an atom that is involved in a denial is a
candidate to be added to a model, DCC checks whether the candidate
model is consistent with the rest of the atoms (including builtins) in
that denial.  Depending on the result of this check, the candidate
atom is added or not.  The modified interpreter performs the following
steps:

\begin{enumerate}
\item The DCC check starts when the interpreter proves that a goal
  \code{G} holds, by invoking the predicate \code{eval_dcc/2}.  If it
  succeeds, \code{G} is added to the model and the evaluation
  continues.  Otherwise, an inconsistency has been detected and
  backtracking takes place.
\item The current implementation of \code{eval_dcc/2} only evaluates
  fully instantiated goals.  Therefore, if the \code{G} is not ground
  (line~\ref{dcc:ground}), \code{eval_dcc/2} succeeds and the
  evaluation continues. This is not a source of unsoundness, as in any
  case the whole set of denials are checked before finally returning a
  model.
\item Otherwise, \code{dcc_rule(G,F_Atoms)} (line~\ref{dcc:dcc})
  retrieves the DCC rules that involve goal \code{G}.
  If there are no DCC rules,  \code{eval_dcc/2} succeeds.
\item Since one atom in the rule (\code{G}) is to be added to the
  model, we need to ensure that not all the rest of the atoms in the
  rule (\code{F_Atoms}) appear in the candidate model \code{In}.  In
  order not to instantiate the model, this is done by checking with
  \code{holds_dcc(F_Atoms,In)} whether \emph{all} the atoms appear in
  \code{In}\footnote{For builtins: checking that they succeed if
    evaluated in the environment of the model.}  and then negating (by
  failure) the calling
  predicate.

\end{enumerate}

Let us consider a program including the denial %
\code{:-p(X), q(X,Y)}.  When the evaluation of \code{p(1)} succeeds,
and before it is added to the tentative model, the interpreter calls
\code{eval_dcc(p(1),[$\dots$])}, where \code{[$\dots$]} is the
current tentative model.
Since \code{p(1)} is ground, \code{pr_dcc_rule(p(1),F_Atoms)}
retrieves in \code{F_Atoms} a list of the atoms that cannot be true in
the model --- in this case, \code{[q(1,Y)]}.
Then, the interpreter checks if the literals in \code{[q(1,Y)]} are
true in the current (tentative) model \code{In}.  If that is the case,
\code{holds_dcc([q(1,Y)],[$\dots$])} succeeds, \code{eval_dcc/2}
fails, and the interpreter backtracks because the denial has been
violated.  Otherwise, the evaluation continues.

It is easy to see that this implementation of dynamic consistency
checking is \emph{complete}, i.e., we do not lose answers: since only
ground goals are checked, there is no risk of instantiating free
variables which could restrict degrees of freedom of the tentative
model and therefore potentially removing solutions.
Furthermore, to ensure \emph{correctness}, we keep the non-monotonic
rule checking that is performed once the tentative model is found.
Note that non-ground goals are at the moment not subject to DCC rules,
but they may be involved in denials, and denials of atoms not needed
to support the query must be checked.

DCC is also used during the execution of the \code{nmr_check} predicate.
As we mentioned before (item~\ref{item:nmr-intro} in
page~\pageref{item:nmr-intro}), denials are compiled into a
synthesized goal, \code{nmr_check}, that is executed after a model has
been generated.  During its execution, DCC rules are actively used to
look for atoms that are introduced and when an inconsistency is
flagged, execution fails and backtracks.

\begin{example}[Cont. Example~\ref{exa:hamiltonian}]

  Let us consider the Hamiltonian program in
  Fig.~\ref{fig:hamiltonian}. As explained above, the compiler
  generates the DCC rules below.  For this example each denial is
  translated into two specialized rules.

\medskip

      \begin{lstlisting}[style=MySCASP]
dcc(vertex(U), [not reachable(U)]).
dcc(not reachable(U), [vertex(U)]).
dcc(chosen(U,W), [U \= V, chosen(V,W)]).
dcc(chosen(V,W), [chosen(U,W), U \= V]).
dcc(chosen(W,U), [U \= V, chosen(W,V)]).
dcc(chosen(W,V), [chosen(W,U), U \= V]).
      \end{lstlisting}

\medskip

  By invoking \code{scasp --dcc hamiltonian.pl graph_4.pl}, s(CASP)
  evaluates the query \code{?-reachable(a)} following a goal-directed
  strategy. Let us refer to the code in Fig.~\ref{fig:hamiltonian} and
  the graph in Fig.~\ref{fig:graph} to explain how the evaluation with
  DCC takes place:
  \begin{enumerate}
  \item The query unifies with the clause in line~\ref{ham:r1} but the goal
    \code{chosen(a,a)} fails because \code{edge(a,a)} does not exist.
  \item From the clause in line~\ref{ham:r2}, \code{chosen(b,a)} is
    added to the tentative model, because no DCC rule succeeds.
    The goal \code{reachable(b)} is then called.
  \item The goal \code{reachable(b)} unifies with the clause in
    line~\ref{ham:r1} and \code{chosen(a,b)} is added, because it is
    consistent with \code{chosen(b,a)}.
  \item As the query succeeds for the model
    \code{\{chosen(b,a), chosen(a,b), reachable(a), reachable(b), $\dots$\}},
    s(CASP) invokes \code{nmr_check}.
    
  \item \code{nmr_check} executes checks for all the denials.  The
    code corresponding to line~\ref{ham:d1} is:

\begin{lstlisting}[style=MySCASP]
chk1 :- forall(U, not chk1_1(U))).
not chk1_1(U) :- not vertex(U).
not chk1_1(U) :- vertex(U), reachable(U).
\end{lstlisting}

    \noindent
    i.e., all vertices (\code{vertex(U)}) must be reachable
    (\code{reachable(U)}). %
    For vertices \code{U = a} and \code{U = b}, \code{reachable(a)}
    and \code{reachable(b)} are already in the model, so there is
    nothing to check.  But for vertex \code{U = c}, \code{reachable(c)} is
    not in the model and therefore \code{reachable(c)} has to be
    invoked while checking the denials. %
  \item From the clause in line~\ref{ham:r1}, \code{chosen(a,c)} is
    selected to be added to the model, but it is discarded by DCC,
    because of the DCC rule %
    \code{dcc(chosen(V,W), [chosen(U,W), U \= V]}), corresponding to
    the denial in line~\ref{ham:d2}. Note that this DCC rule is
    instantiated to
    \code{dcc(chosen(a,c), [c \= b, chosen(a,b)]}
    and the literal \code{chosen(a,b)} is already in the model.
    
  \item The evaluation backtracks and continues the search using another
    edge.
  \end{enumerate}

  The denial in line~\ref{ham:d1} makes the interpreter to select
  edges to reach all vertices.  The interleaving of the dynamic
  consistency checking prunes the search, which, as shown in
  Section~\ref{sec:evaluation}, improves performance.
  
\end{example}

\section{Evaluation}
\label{sec:evaluation}

In this section we compare the performance of s(CASP) with and without
DCC using a macOS 11.5.2 Intel Core i7 at 2.6GHz. We use s(CASP)
version 0.21.10.09 available at
\url{https://gitlab.software.imdea.org/ciao-lang/scasp} and, as
mentioned before, all the benchmarks used or mentioned in this paper
are available at
\url{http://platon.etsii.urjc.es/~jarias/papers/dcc-padl21}.

\begin{table}[t]\renewcommand{\arraystretch}{1.25}\setlength{\tabcolsep}{10pt}
  \centering
  \caption{Performance comparison:  s(CASP) and s(CASP)+DCC.}
  \label{tab:eval}
  \begin{tabular}{lrrr}
    \toprule
    & Speedup& s(CASP) & s(CASP)+DCC  \\
    \midrule                         
    Hamiltonian (4 vertices)         & 10.0    & 11.985       & \textbf{1.196}      \\
    Hamiltonian (7 vertices)         & 41.1    & 134.460      & \textbf{3.191}      \\
    n\_queens (\code|n=4|)           & 4.3     & 8.147        & \textbf{1.910}      \\
    n\_queens (\code|n=5|)           & 4.9     & 92.756       & \textbf{18.786}     \\
    n\_queens (\code|n=6|)           & 90.8    & 1362.840     & \textbf{15.001}     \\
    n\_queens\_attack (\code|n=6|)   & 1.0     & 77.039       & \textbf{76.827}     \\
    \bottomrule
  \end{tabular}
\end{table}

Table~\ref{tab:eval} shows the results of the performance comparison
(in seconds), and the speedup of s(CASP) with DCC w.r.t.\ s(CASP).

First, we evaluate the Hamiltonian path problem
(Example~\ref{exa:hamiltonian}) using the encoding in
Fig.~\ref{fig:hamiltonian} (available at \myurl{hamiltonian.pl}) and
the graph with 4 vertices in Fig.~\ref{fig:graph} (available at
\myurl{graph_4.pl}).
We see that s(CASP) with DCC obtains a speedup of
\textbf{10.0$\times$}.
When the size of the graph is increased by adding three
vertices\footnote{The graph with seven vertices
  is available at \myurl{graph_7.pl}.} 
we obtain a speedup of \textbf{41.1$\times$}.
 
We also evaluated the performance of s(CASP) with DCC with the
well-known $n$-queens problem, using two different versions.  These
examples are especially interesting, as they have no finite grounding
and thus cannot be run by other implementations of the stable model
semantics.  In particular, the size of the board is not fixed by the
programs and therefore the same code can be used to find solutions to
several board sizes.

The first version, \code{n_queens} (available in
Appendix~\ref{sec:encod-coden_q} and at \myurl{n_queens.pl}), uses
denials to discard solutions where two queens attack each other. 
The speedup obtained by s(CASP) with DCC ranges from
\textbf{4.3$\times$} (for $n = 4$) to \textbf{90.8$\times$} (for
$n = 6$).

The second version is the one presented
in~\cite[Pag.~37]{marple2012:goal-directed-asp} (available in
Appendix~\ref{sec:encod-coden_q-1} and at \myurl{n_queens_attack.pl}).
In this case, the predicate \code{attack/3} is used to check whether a
candidate queen attacks any previously selected queen.  This version
does not have denials, as the check is done as part of the
computation, and therefore DCC checks are not useful.

Two interesting conclusions can be drawn from the numbers in the
table:

\begin{itemize}
\item The execution time of \code{n_queens_attack} does not
  significantly change using or not DCC, which supports our assumption
  that the overhead of using DCC checks when they are not needed is
  negligible.
\item On the other hand, the DCC-enabled execution for the version
  with denials (n\_queens, column ``s(CASP)+DCC'') is faster than the
  version without denials (n\_queens\_attack, column ``s(CASP)'') for
  a factor of \textbf{4.1$\times$}.  This can be attributed to the
  runtime being sophisticated enough to perform checks earlier and
  more efficiently than the hand-crafted code, even with the current,
  preliminary implementation.
\end{itemize}

\begin{table}[t]\renewcommand{\arraystretch}{1.25}\setlength{\tabcolsep}{5pt}
  \centering
  \caption{Models generated and/or discarded: s(CASP)
    vs. s(CASP)+DCC.}
  \label{tab:models_a}
  \begin{tabular}{lrp{.1cm}rrr}
    \toprule
    &  \multirow{2}{1.3cm}{\hfill \#models\\\hfill returned} & & \multicolumn{2}{c}{\#models discarded} & \multirow{2}{1.4cm}{\hfill \#DCC\\\hfill detected} \\ \cline{4-5}
    &   & & s(CASP) & s(CASP)+DCC &  \\
    \midrule                         
    Hamiltonian (4 vertices)          & 3  & & 7     & 7  & 52  \\ 
    Hamiltonian (7 vertices)          & 1  & & 13    & 13 & 34  \\ 
    n\_queens (\code|n=4|)            & 2  & & 253   & 0  & 44  \\  
    n\_queens (\code|n=5|)            & 10 & & 3116  & 0  & 167 \\  
    n\_queens (\code|n=6|)            & 4  & & 46652 & 0  & 742 \\  

    \bottomrule
  \end{tabular}
\end{table}

Table~\ref{tab:models_a}  sheds some additional
light on the effectiveness of DCC.  It contains, for the same
benchmarks as Table~\ref{tab:eval}, how many models were returned, how
many candidate models were discarded by \code{nmr_check} after they were
generated (column ``\#models discarded -- s(CASP)''), how many
(partial) candidate models were discarded using DCCs (column
``\#models discarded -- s(CASP) + DCC'') and how many times the dynamic consistency checking detects an inconsistency and backtracks
(column ``\#DCC detected'').

Let us first focus on the n\_queens
benchmark.\footnote{n\_queens\_attack does not have denials, hence we
  do not include it in this table.}  As the size of the board grows,
the number of models that are completely generated and discarded by
s(CASP) without using DCCs grows exponentially; this is of course the
reason why its execution time also increases very quickly.  If we look
at the column ``\#models discarded -- s(CASP) + DCC'' we see that,
when DCC is active, none of final models is discarded by
\code{nmr_check}.  That means that all models not consistent with the
denials have been removed early, while they were being built.  We also
see that the number of times that DCC rules were activated is much
smaller than the number of times that \code{nmr_check} was executed
--- early %
pruning made it possible to avoid attempting to generate many other
candidate models.  On top of that, executing DCC rules is done
directly in Prolog, while \code{nmr_check} is executed using the 
s(CASP) metainterpreter, and the former is considerably faster than
the latter.  This adds to the advantage of early pruning.

The Hamiltonian path benchmark is different and very interesting.  The
number of models discarded by \code{nmr_check} is the same regardless
of whether DCC is activated or not.  That means that the DCC could not
detect inconsistencies in candidate models.  In this case the
advantage of using DCC comes from applying it when \code{nmr_check} is
invoked to ensure that the final model is consistent with the denials.
\code{nmr_check} is executed as a piece of (synthesized) code by the
metainterpreter.  The denials in the Hamiltonian path not only check,
but also generate new atoms which are checked by the DCC.  This
accelerates the execution of \code{nmr_check}, making it fail earlier,
and it is the cause of the speedup of the Hamiltonian path benchmark.

\section{Conclusions}
\label{sec:conclusions}

In this paper, we have reported on a preliminary design and
implementation of Dynamic Consistency Checking (DCC), a technique that
anticipates the consistency evaluation of tentative models in s(CASP),
a goal-directed (predicate) Constraint Answer Set Programming.
This technique translates the denials
to check them as early as possible rather than when a full model is
found for a given query.
Its ability to detect inconsistencies before a literal is added to the
tentative model greatly increases the performance.
with respect with executions without DCC.  Early denial checking can
also beat programs that use auxiliary predicates explicitly called
from the user code to check for inconsistencies.

The current DCC implementation can still be improved, in particular to
properly handle constraints by reducing the domain of constrained
variables:
checking denials using literals with constrained variables has to keep
track of the domains of the variables, while avoiding introducing
non-determinism when conflicting values are removed from the domain of
the variables.

%

\newpage
\appendix

\section{Encoding of \code{n_queens.pl}}
\label{sec:encod-coden_q}

\begin{lstlisting}[style=MySCASP]
% N-queens program.
nqueens(N, Q) :-
    nqueens_(N, N, [], Q).

% Pick queens one at a time.
nqueens_(X, N, Qi, Qo) :-
    X > 0,
    pickqueen(X, Y, N),
    X1 is X - 1,
    nqueens_(X1, N, [queen(X, Y) | Qi], Qo).
nqueens_(0, _, Q, Q).

% pick a queen for row X.
pickqueen(X, Y, Y) :-
    Y > 0,
    queen(X, Y).
pickqueen(X, Y, N) :-
    N > 1,
    N1 is N - 1,
    pickqueen(X, Y, N1).

queen(X, Y) :- not neg_q(X, Y).
neg_q(X, Y) :- not queen(X, Y).
#show queen/2.

% Test
:- queen(I,J1), queen(I,J2), J1 \= J2.
:- queen(I1,J), queen(I2,J), I1 \= I2.
:- queen(I,J), queen(II,JJ), I\=II, J\=JJ, T1 is I+J, T2 is II+JJ, T1=T2.
:- queen(I,J), queen(II,JJ), I\=II, J\=JJ, T1 is I-J, T2 is II-JJ, T1=T2.
\end{lstlisting}

\newpage

\section{Encoding of \code{n_queens_attack.pl}}
\label{sec:encod-coden_q-1}

\begin{lstlisting}[style=MySCASP]
nqueens(N, Q) :-
    nqueens_(N, N, [], Q).

% Pick queens one at a time and test against all previous queens.
nqueens_(X, N, Qi, Qo) :-
    X > 0,
    pickqueen(X, Y, N),
    not attack(X, Y, Qi),
    X1 is X - 1,
    nqueens_(X1, N, [q(X, Y) | Qi], Qo).
nqueens_(0, _, Q, Q).

% pick a queen for row X.
pickqueen(X, Y, Y) :-
    Y > 0,
    q(X, Y).
pickqueen(X, Y, N) :-
    N > 1,
    N1 is N - 1,
    pickqueen(X, Y, N1).

% check if a queen can attack any previously selected queen.
attack(X, _, [q(X, _) | _]).        % same row
attack(_, Y, [q(_, Y) | _]).        % same col
attack(X, Y, [q(X2, Y2) | _]) :-    % same diagonal 1
    T1 is X + Y, T2 is X2 + Y2, T1 = T2.
attack(X, Y, [q(X2, Y2) | _]) :-    % same diagonal 2
    T1 is X - Y, T2 is X2 - Y2, T1 = T2.
attack(X, Y, [_ | T]) :- 
    attack(X, Y, T).

q(X, Y) :- not neg_q(X, Y).
neg_q(X, Y) :- not q(X, Y).
#show q/2.
\end{lstlisting}

\end{document}